# An Application of Neutrosophic Sets to Decision Making

**Michael Gr. Voskoglou[1] ***

[1] School of Engineering, University of Peloponnese (ex Graduate TEI of Western Greece), 26334 Patras, Geece; voskoglou@teiwest.gr
* Correspondence: mvoskoglou@gmail.com

**Abstract:** Frequently in real life situations decision making takes place under fuzzy conditions, because the corresponding goals and/or the existing constraints are not clearly defined. Maji et al. introduced in 2002 a method of parametric decision making using soft sets as tools and representing their tabular form as a binary matrix. As we explain here, however, in cases where some or all of the parameters used for the characterization of the elements of the universal set are of fuzzy texture, their method does not give always the best decision making solution. In order to tackle this problem, we modified in earlier works the method of Maji et al. by replacing the binary elements in the tabular form of the corresponding soft set either by grey numbers or by triangular fuzzy numbers. In this work, in order to tackle more efficiently cases in which the decision maker has doubts even about the correctness of the fuzzy/qualitative characterizations assigned to some or all of the elements of the universal set, we replace the binary elements of the tabular form by neutrosophic triplets. Our new, neutrosophic decision making method is illustrated by an application concerning the choice of a new player by a soccer club.

## 1. Introduction

*Decision Making (DM)* is a fundamental process in a great spectrum of human activities and many books have been written about it, helping decision makers to make smarter choices easier and quicker; e.g. see [1, 2], etc. Frequently in real life situations, however, DM takes place under fuzzy conditions, since the corresponding goals and/or the existing constraints are not clearly defined. Several methods have been also proposed for successful DM in such cases, e.g. [3-5], etc.

Maji et al. introduced in 2002 a method of parametric DM using *soft sets (SS)* as tools and representing their tabular form as a binary matrix [6]. When some or all of the parameters used for the characterization of the elements of the set of the discourse (houses in their example) are of fuzzy texture (beautiful and cheap in their example), however, their method does not always give the best solution to the corresponding DM problem. This happens, because they replace the parameters in the tabular form of the corresponding SS with the binary elements 0, 1. In other words, their method, although it starts with a fuzzy framework (SS), it continues by using bivalent logic for obtaining the required decision (beautiful or not beautiful and cheap or not cheap in their example)! In order to tackle this problem, we modified in earlier works the method of Maji et al. by using either *triangular fuzzy numbers (TFNs)* [7] or *grey numbers (GNs)* – see [8], or [9] (section 5.3) - instead of the binary elements in the tabular form of the corresponding SS.

In reality, however, cases appear in which the decision maker has doubts about the correctness of the fuzzy/qualitative characterizations assigned to some or all of the elements of the set of the discourse (e.g. very beautiful, rather beautiful, etc.). In order to study more efficiently the DM process in the previous cases, we introduce here

*neutrosophic sets (NSs)* and we replace the binary elements of the method of Maji et al. by *neutrosophic triplets*. The rest of the paper is organized as follows: Section 2 contains the necessary background about NSs, SSs and GNs needed for the understanding of the paper. The DM method of Maji et al., our modification using GNs, as well as the new neutrosophic DM method are developed in section 3, illustrated by examples concerning the choice of a new player by a soccer club. The article closes with a discussion on the results obtained in it, including some hints for future research, and the final conclusions, presented in section 4.

## 2. Mathematical Background

*2.1 Fuzzy Sets and Fuzzy Logic*

Zadeh extended in 1965 the concept of crisp set to the concept of FS [10] on the purpose of dealing with the existing in real life partial truths (e.g. rather good, almost true, etc.) and of expressing mathematically definitions with no clear boundaries (e.g. high mountains, clever people, etc.) This was succeeded by replacing the characteristic function by the *membership function*, which maps each element of the universal set U to the unit interval [0, 1]. In fact, if A is the corresponding FS in U and m: U → [0, 1] is its membership function, m(x) is called the *membership degree* of x in A, for all elements x in U. The closer m(x) to 1, the better x satisfies the characteristic property of A. And, although the FS A is typically defined as the set of all ordered pairs of the form (x. m(x)), ∀ x∈U, for reasons of simplicity many authors identify A with its membership function.

Most of notions and operations on crisp sets are extended in a natural way to FSs (e.g. see [11]). Based on the concept of FS, Zadeh introduced the infinite-valued *fuzzy logic (FL)*, in which the truth values are modelled by numbers in the unit interval [12]. FL does not oppose the traditional bivalent logic of Aristole (384-322 BC), which used to be for many centuries the basic tool of human reasoning being "responsible" for the growth of science and human civilization all this time; on the contrary it completes and extends it [13]

In a later stage, when membership functions were reinterpreted as *possibility* distributions, FSs and FL were used to embrace *uncertainty* modelling [14, 15]. The frequently appearing in real life uncertainty is due to several reasons, like randomness, imprecise or incomplete data, vague information, etc. *Probability* theory has been proved sufficient to tackle only the cases of uncertainty which are due to randomness [16]. Starting from Zadeh's FSs, however, several theories have been developed during the last years on the purpose of tackling more effectively all the forms of the existing uncertainty. The main among those theories are briefly reviewed in [17]. In the present paper we are going to use elements from the theories of NSs, SSs and GNs needed for its understanding, which are exposed below.

*2.2 Neutrosophic Sets*

Atanassov in 1986, considered in addition to Zadeh's membership degree the degree of *non-membership* and extended FS to the notion of *intuitionistic FS (IFS)* [18]. Smarandache in 1995, inspired by the frequently appearing in real life neutralities - like <friend, neutral, enemy>, <win, draw, defeat>, <high, medium, short>, etc. - generalized IFS to the concept of NS by adding the degree of *indeterminacy* or *neutrality* [19]. The word "neutrosophy" is a synthesis of the word "neutral´ and the Greek word "sophia" (wisdom) and means "the knowledge of neutral thought". The simplest form of a NS is defined as follows:

**Definition 1:** A *single valued NS (SVNS)* A in the universe U is of the form

A = {(x,T(x),I(x),F(x)): x∈U, T(x),I(x),F(x)∈[0,1], 0≤T(x)+I(x)+F(x)≤3}   (1)

In equation (1) T(x), I(x), F(x) are the degrees of *truth* (or membership), *indeterminacy* (or neutrality) *and falsity* (or non-membership) of x in A respectively, called the *neutrosophic components* of x. For simplicity, we write A<T, I, F>.

Indeterminacy is defined to be in general everything that exists between the opposites of truth and falsity [20].

**Example 1:** Let U be the set of the players of a soccer club and let A be the SVNS of the good players of the club. Then each player x is characterized by a *neutrosophic triplet* (t, i, f) with respect to A, with t, i, f in [0, 1]. For example, x(0.7, 0.1, 0.4) ∈ A means that the coach of the club is 70% sure that x is a good player, but at the same time he has a 10% doubt about it and a 40% belief that x is not a good player. In particular, x (0, 1, 0) ∈ A means that the coach does not know absolutely nothing about player x (new player).

If the sum T(x) + I(x) + F(x) < 1, then it leaves room for incomplete information about x, if it is equal to 1 for complete information and if it is >1 for *inconsistent* (i.e. contradiction tolerant) information about x. A SVNS may contain simultaneously elements leaving room to all the previous types of information. All notions and operations defined on FSs are naturally extended to SVNSs [21].

Summation of neutrosophic triplets is equivalent to the union of NSs. That is why the neutrosophic summation and implicitly its extension to neutrosophic scalar multiplication can be defined in many ways, equivalently to the known in the literature neutrosophic union operators [22]. For the needs of the present work, writing the elements of a SVNS A in the form of neutrosophic triplets and considering them simply as <u>ordered triplets</u> we define addition and scalar product as follows:

**Definition 2:** Let $(t_1, i_1, f_1)$, $(t_2, i_2, f_2)$ be in A and let k be a positive number. Then:

- The *sum* $(t_1, i_1, f_1) + (t_2, i_2, f_2) = (t_1+ t_2,\ i_{1+}\ i_2,\ f_{1+}\ f_2)$   (2)
- The *scalar product* $k(t_1, i_1, f_1) = (kt_1,\ k\ i_1,\ kf_1)$   (3)

**Remark 1:** The summation and scalar product of the elements of a SVNS A with respect to Definition 2 <u>need not</u> be a closed operation in A, since it may happen that $(t_1+ t_2)+(i_{1+}\ i_2)+(f_{1+}\ f_2)>3$ or $kt_1+k\ i_1+kf_1>3$. With the help of Definition 2, however, one can define <u>in A</u> the *mean value* of a finite number of elements of A as follows:

**Definition 3:** Let A be a SVNS and let $(t_1, i_1, f_1)$, $(t_2, i_2, f_2)$, …., $(t_k, i_k, f_k)$ be a finite number of elements of A. Assume that $(t_i, i_i, f_i)$ appears $n_i$ times in an application, i = 1,2,…., k. Set $n = n_1+n_2+….+n_k$. Then the *mean value* of all these elements of A is defined to be <u>the element of A</u>

$$(t_m, i_m, f_m) = \frac{1}{n}[n_1(t_1, i_1, f_1)+n_2(t_2, i_2, f_2)+….+n_k(t_k, i_k, f_k)]  \quad (4)$$

*2.3 Soft Sets*

A disadvantage of FSs and of all their extensions involving membership degrees (like IFSs, NSs, etc.), is that there is not any exact rule for defining properly the corresponding membership functions. The methods used for this are usually statistical or intuitive/empirical.  Moreover, the definition of the membership function is not unique depending on the "signals" that each observer receives from the environment. For example, defining the FS of "young people" one could consider as young all persons aged less than 30 years and another one all persons aged less than 40 years. As a result the second observer will assign membership degree 1 to all people aged between 30 and 40 years, whereas the first one will assign to them membership degrees less than 1. Analogous differences are logically expected to appear to the membership degrees of all the other ages. In other words, the only restriction for the definition of the membership function is that it must be compatible to the common sense; otherwise the resulting FS does not give a creditable description of the corresponding real situation. This could happen, for instance, if in the previous example people aged more than 70 years possessed membership degrees ≥0.5.

For overpassing this problem, the concept of *interval-valued FS (IVFS)* was introduced in 1975. An IVFS is defined by mapping the universe U to the set of the closed subintervals of [0, 1] [23]. Other related to FS theories were also developed in which the definition of a membership function is not necessary (*grey systems and numbers* [24]), or it is overpassed by using a pair of crisp sets giving the lower and upper bound of the original set (*rough sets [25]*). Molodstov, introduced in 1999 the concept of SS for tackling the uncertainty in a parametric manner, not needing, therefore, the definition of a membership function [26]. Namely, a SS is defined as follows:

**Definition 4:** Let E be a set of parameters, let A be a subset of E, and let f be a map from A into the power set P(U) of the universe U. Then the SS (f, A) in U is defined as the set of the ordered pairs

$$(f, A) = \{(e, f(e)): e \in A\} \quad (5)$$

In other words, a SS is a parametrized family of subsets of U. The name "soft" was given due to the fact that the form of (f, A) depends on the parameters of A. For each $e \in A$, its image f(e) in P(U) is called the *value set* of e in (f, A), while f is called the *approximation function* of (f, A).

**Example 2:** Let U= {$H_1$, $H_2$, $H_3$} be a set of houses and let E = {$e_1$, $e_2$, $e_3$} be the set of parameters $e_1$=cheap, $e_2$=beautiful and $e_3$= expensive. Let us further assume that $H_1$, $H_2$ are cheap, $H_3$ is expensive and $H_2$, $H_3$ are beautiful houses. Then, a map f: E → P(U) is defined by f($e_1$)={$H_1$, $H_2$}, f($e_2$)={$H_2$, $H_3$} and f($e_3$)={$H_3$}. Therefore, the SS (f, E) in U is the set of the ordered pairs

$$(f, E) = \{(e_1, \{H_1, H_2\}), (e_2, \{H_2, H_3\}), (e_3, \{H_3\})\} \quad (6)$$

Maji et al. [6] introduced a *tabular representation* of SSs in the form of a binary matrix in order to be stored easily in a computer's memory. For instance, the tabular representation of the soft set (f, E) of the previous example is given by Table 1.

**Table 1.** Tabular representation of the SS of Example 2

|  | $e_1$ | $e_2$ | $e_3$ |
|---|---|---|---|
| $H_1$ | 1 | 0 | 0 |
| $H_2$ | 1 | 1 | 0 |
| $H_3$ | 0 | 1 | 1 |

A FS in U with membership function y = m(x) is a SS in U of the form (f, [0, 1]), where f(α)={$x \in U$: m(x)≥α} is the corresponding *a-cut* of the FS, for each α in [0, 1].

Consequently the concept of SS is a generalization of the concept of FS. All notions and operations defined on FSs are extended in a natural way to SSs [27].

*2.4 Grey Numbers*

The theory of *grey systems* [24] introduces an alternative way for managing the uncertainty in case of approximate data. A grey system is understood to be any system which lacks information, such as structure message, operation mechanism or/and behavior document.

Closed real intervals, are used for performing the necessary calculations in grey systems. In fact, a closed real interval [x, y] could be considered as representing a real number T, termed as a *grey number (GN)*, whose exact value in [x, y] is unknown. We write then T ∈ [x, y]. A GN T, however, is frequently accompanied by a *whitenization function* f: [x, y] → [0, 1], such that, if f(a) approaches 1, then a in [x, y] approaches the unknown value of T. If no whitenization function is defined, it is logical to consider as a representative crisp approximation of the GN A the real number

$$V(A) = \frac{x+y}{2} \qquad (7)$$

The arithmetic operations on GNs are introduced with the help of the known arithmetic of the real intervals [28]. In this work we are going to make use only of the addition of GNs and of the scalar multiplication of a GN with a positive number, which are defined as follows:

**Definition 5:** Let A ∈ [$x_1$, $y_1$], B ∈ [$x_2$, $y_2$] be two GNs and let k be a positive number. Then:

- The sum: A+B is the GN A+B ∈ [$x_1+y_1$, $x_2+y_2$]   (8)
- The scalar product kA is the GN kA ∈ [$kx_1$, $ky_1$]   (9)

**3. The Parametric Assessment Method**

The parametric assessment method of Maji et al. [6], our modification using GNs [8, 9] and our new method using NSs are developed in this section through suitable examples concerning the choice of a new player by a soccer club.

*3.1 The Method of Maji et al.*

Let us consider the following example:

**Example 3:** A soccer club wants to choose a new player among 6 candidates, say $P_1$, $P_2$, $P_3$, $P_4$, $P_5$ and $P_6$. The desired qualifications of the new player are to be fast, younger than 30 years, higher than 1.70 m and experienced. Assume that $P_1$, $P_2$, $P_6$ are the fast players, $P_2$, $P_3$, $P_5$, $P_6$ are the players being younger than 30 years, $P_3$, $P_5$ are the players with heights greater than 1.70 m and $P_4$ is the unique experienced player. Which is the best choice for the club?

*Solution:* Let U be the set of the 6 candidate players. Consider the parameters $e_1$=fast, $e_2$=younger than 30 years, $e_3$=higher than 1.70 m, $e_4$=experienced and set E = {$e_1$, $e_2$, $e_3$, $e_4$}. Define the map f: E → P(U) by f($e_1$) = {$P_1$, $P_2$, $P_6$}, f($e_2$) = {$P_2$, $P_3$, $P_5$, $P_6$}, f($e_3$) = {$P_3$, $P_5$} and f($e_4$) = {$P_4$}. Then the tabular form of the SS (f, E) is shown in Table 2.

**Table 2.** Tabular representation of the SS of Example 3

|    | e₁ | e₂ | e₃ | e₄ |
|----|----|----|----|----|
| P₁ | 1  | 0  | 0  | 0  |
| P₂ | 1  | 1  | 0  | 0  |
| P₃ | 0  | 1  | 1  | 0  |
| P₄ | 0  | 0  | 0  | 1  |
| P₅ | 0  | 1  | 1  | 0  |
| P₆ | 1  | 1  | 0  | 0  |

The *choice value* of each player is calculated by adding the binary elements of the row of Table 2 in which he belongs. The players $P_1$ and $P_4$, therefore, have choice value 1 and all the others have choice value 2. Consequently, the right decision is to choose one of the players $P_2$, $P_3$, $P_5$, and $P_6$.

*3.2 Our Method Using GNs*

The previous decision, obtained by the method of Maji et al., is obviously not very helpful for the soccer club. Observe, however, that the parameters $e_1$ and $e_4$, in contrast to $e_2$ and $e_3$, have not a bivalent texture. A player, for example, could not be very fast, but quite fast, or rather experienced, etc. This inspired us to characterize the qualifications of the players with respect to the parameters $e_1$ and $e_4$ in the tabular matrix of the previous SS (f, E) by using the linguistic grades A=excellent, B=very good, C=good, D=mediocre and F= not satisfactory instead of the binary element 0. To show how one works in this case for making the right decision, let us modify Example 3 as follows:

**Example 4:** Reconsider Example 3 and assume that the technical manager of the soccer club, after a more careful inspection of the qualifications of the 6 candidate players, decided to use the following Table 3 instead of Table 2 for the DM process. Which is the best choice for the club in this case?

**Table 3.** Tabular representation of the SS of Example 4

|    | e₁ | e₂ | e₃ | e₄ |
|----|----|----|----|----|
| P₁ | 1  | 0  | 0  | C  |
| P₂ | 1  | 1  | 0  | F  |
| P₃ | C  | 1  | 1  | C  |
| P₄ | D  | 0  | 0  | 1  |
| P₅ | D  | 1  | 1  | C  |
| P₆ | 1  | 1  | 0  | D  |

*Solution:* Assign to each of the qualitative grades A, B, C, D, F a GN, denoted for simplicity by the same letter, as follows: A = [0.85, 1], B = [0.75, 0.84], C= [0.6, 0.74], D = [0.5, 0.59], F = [0, 0.49]. From Table 3 then, one calculates, with the help of formulas (7), (8) and (9), the choice value of each player in the following way. $P_1$:

$1+V(C) = 1+\frac{0.6+0.74}{2} = 1.67$, $P_2$: $2+V(F) = 2.245$, $P_3$: $2+V(2C) = 3.34$, $P_4$: $1+V(D)= 1.545$, $P_5$: $2+V(D+C) = 3.215$, $P_6$: $2+V(D)=2.545$. The right decision, therefore, is to choose the player $P_3$.

**Remark 2:** The choices of the qualitative grades A, B, C, D, F, as well as of the intervals for translating them in the numerical scale 0-1, correspond to generally accepted standards. The decision maker, however, could use, with respect to his/her goals, more or less qualitative grades (e.g. by adding E between D and F, etc.) and could also change the corresponding intervals (e.g. by setting A= [0.9, 1], B = [0.8, 0.89], C= [0.7, 0.79], D = [0.6, 0.69], F = [0, 0.59], or otherwise). Such changes, however, does not affect the generality of our method

*3.3. The Neutrosophic DM Method*

As it was already mentioned in our Introduction, DM situations appear frequently in reality, in which the decision maker has doubts about the correctness of the fuzzy/qualitative grades assigned to some or all of the elements of the set of the discourse. In such cases, the best way to perform the DM process is to use NSs. In Example 4, for instance, considering the set U of the 6 candidate players as the universal set, the decision maker could define in U the NSs of the fast and of the experienced players by assigning the suitable neutrosophic triplets to each player of U. In order to have complete information, we should have t+i+f = 1, for each triplet (t, i, f). Then, he/she could continue the DM process by replacing in Table 3 the qualitative grades by the corresponding neutrosophic triplets and the binary elements 0, 1 by the neutrosophic triplets (0, 0, 1) and (1, 0, 0) respectively. This process will be illustrated by the following Example 5.

**Example 5:** Reconsider Example 4 and assume that the technical manager of the soccer club, being not sure about the qualitative grades assigned to each of the 6 candidate players, he decided to proceed by replacing them by neutrosophic triplets, in the way that we have previously described. As a result, the tabular matrix of the DM process took the form of the following Table 4. Which is the best decision for the club in this case?

**Table4.** Tabular representation of the SS of Example 5

|       | $e_1$          | $e_2$     | $e_3$     | $e_4$           |
|-------|----------------|-----------|-----------|-----------------|
| $P_1$ | (1, 0, 0)      | (0, 0, 1) | (0, 0, 1) | (0.6, 0.3, 0.1) |
| $P_2$ | (1, 0, 0)      | (1, 0, 0) | (0, 0, 1) | (0.2, 0.2, 0.6) |
| $P_3$ | (0.5, 0.4, 0.1)| (1, 0, 0) | (1, 0, 0) | (0.6, 0.2, 0.2) |
| $P_4$ | (0.5, 0.2, 0.3)| (0, 0, 1) | (0, 0, 1) | (1, 0, 0)       |
| $P_5$ | (0.5, 0.1, 0.4)| (1, 0, 0) | (1, 0, 0) | (0.6, 0.3, 0.1) |
| $P_6$ | (1, 0, 0)      | (1, 0, 0) | (0, 0, 1) | (0.4, 0.4, 0.2) |

*Solution:* The choice value of each player in this case is defined to be the mean value of the neutrosophic triplets of the line of Table 4 in which he belongs. Thus, by equation (4), the choice value of $P_1$ is equal to $\frac{1}{4}[(1, 0, 0)+2(0, 0, 1)+(0.6, 0.3, 0.1)] = \frac{1}{4}(1.6, 0.3, 2.1) = (0.4, 0.075, 0.525)$. In the same way one finds that the choice values of $P_2$,

$P_3$, $P_4$, $P_5$ and $P_6$ are (0.55, 0.005, 0.4), (0.775, 0.15, 0.075), (0.375, 0.05, 0.575), (0.775, 0.1, 0.125) and (0.6, 0.1, 0.3) respectively.

In this case the club's technical manager could use either an optimistic criterion by choosing the player with the greatest truth degree, or a conservative criterion by choosing the player with the lower falsity degree. Consequently, using the optimistic criterion he must choose one of the players $P_3$ and $P_5$, whereas using the conservative criterion he must choose the player $P_3$. A combination of the two criteria leads to the final choice of player $P_3$. Observe, however, that, since the indeterminacy degree of $P_3$ is 0.15 and of $P_5$ is 0.1, there is a slightly greater doubt about the qualifications of $P_3$ with respect to the qualifications of $P_5$. In other words, the choice of $P_3$ is connected with a slightly greater risk. In final analysis, therefore, all the neutrosophic components assigned to each player give useful information about his qualifications.

## 4. Discussion and Conclusions

In this paper a parametric DM method of hybrid character was presented illustrated by suitable examples about the choice of a new player from a football club. The whole discussion performed leads to the following conclusions:

- The parametric DM method is based on the introduction of a set E of parameters characterizing the elements of the set U of all possible solutions of the corresponding DM problem, the definition with respect to E of a suitable SS in U and the use of its tabular representation T as a tool for the DM process.
- When all the parameters of E are of bivalent texture (yes or no), then T takes the form of a binary matrix, wherefrom the decision maker calculates the choice value of each element of U by adding the binary elements of the row of T in which this element appears.
- When some or all of the parameters of E are of fuzzy texture and the decision maker has no doubts about the qualitative characterizations assigned to the elements of U with respect to these parameters, then the binary elements of T corresponding to the fuzzy parameters are replaced by suitable GNs and the choice value of each element of U is calculated by adding the remaining in the corresponding row of T binary elements and the representative real values of the GNs appearing in it.
- When some or all of the parameters of E are of fuzzy texture and the decision maker do has doubts about the qualitative characterizations assigned to the elements of U with respect to these parameters, then each parameter of E is expressed in the form of a NS in U and the binary elements of T are replaced by the corresponding neutrosophic triplets. In this case, the choice value of each element of U is obtained by calculating the mean value of the neutrosophic triplets of the row of T in which this element appears.

As it has been already mentioned in section 2.1 of this paper, several theories extending / generalizing Zadeh's FSs have been developed during the last years on the purpose of tackling more effectively the existing in real life uncertainty. None of these theories, however, can tackle efficiently alone all the forms of the existing uncertainty, but the combination of all of them provides an adequate framework towards this target.

The results obtained in this and earlier works of the present author give strong indications that hybrid methods applied to several situations in fuzzy environments could give better results, not only for DM, as it happened here, but also for assessment - e.g. see [9] (section 5.2) – and probably for many other topics. This is, therefore, a promising area for further research.